\documentclass{article}

\usepackage{arxiv}

\usepackage{amsmath,amsthm}
\usepackage[utf8]{inputenc} 
\usepackage[T1]{fontenc}    
\usepackage{hyperref}       
\usepackage{url}            
\usepackage{microtype}      
\usepackage{cleveref}       
\usepackage{lipsum}         
\usepackage{graphicx}
\usepackage[numbers]{natbib}
\usepackage{doi}

\usepackage{graphicx}
\usepackage{subcaption}
\usepackage{booktabs}
\usepackage{nicefrac}
\usepackage{multirow}
\usepackage{amsfonts}
\usepackage{lipsum}
\usepackage{float}
\usepackage{algorithm}
\usepackage{algorithmic}
\usepackage[acronym]{glossaries}
\usepackage{tikzit}
\usepackage{wrapfig}
\usepackage{nicefrac}
\usepackage{fontawesome}
\graphicspath{{./Figures}}

\title{Federated Dataset Dictionary Learning for Multi-Source Domain Adaptation}

\date{}

\newif\ifuniqueAffiliation
\uniqueAffiliationtrue

\ifuniqueAffiliation 
\author{Fabiola Espinoza Castellon\footnotemark[1]\\
        CEA, List\\
        Université Paris-Saclay\\
        F-91120 Palaiseau, France\\
	\And
	Eduardo Fernandes Montesuma\footnotemark[1] \\
        CEA, List\\
        Université Paris-Saclay\\
        F-91120 Palaiseau, France\\
	\And
	Fred Ngol\`e Mboula \\
        CEA, List\\
        Université Paris-Saclay\\
        F-91120 Palaiseau, France\\
	\And
	Aur\'elien Mayoue\\
        CEA, List\\
        Université Paris-Saclay\\
        F-91120 Palaiseau, France\\
        \And
	Antoine Souloumiac\\
        CEA, List\\
        Université Paris-Saclay\\
        F-91120 Palaiseau, France\\
        \And
	Cédric Gouy-Pailler\\
        CEA, List\\
        Université Paris-Saclay\\
        F-91120 Palaiseau, France\\
}
\else
\usepackage{authblk}

\newbox{\orcid}\sbox{\orcid}{\includegraphics[scale=0.06]{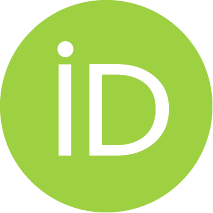}} 
\author[1]{%
	\href{https://orcid.org/0000-0000-0000-0000}{\usebox{\orcid}\hspace{1mm}David S.~Hippocampus\thanks{\texttt{hippo@cs.cranberry-lemon.edu}}}%
}
\author[1,2]{%
	\href{https://orcid.org/0000-0000-0000-0000}{\usebox{\orcid}\hspace{1mm}Elias D.~Striatum\thanks{\texttt{stariate@ee.mount-sheikh.edu}}}%
}
\affil[1]{Department of Computer Science, Cranberry-Lemon University, Pittsburgh, PA 15213}
\affil[2]{Department of Electrical Engineering, Mount-Sheikh University, Santa Narimana, Levand}
\fi

\renewcommand{\shorttitle}{Federated Dataset Dictionary Learning}

\hypersetup{
pdftitle={Federated Dataset Dictionary Learning for Multi-Source Domain Adaptation},
pdfsubject={cs.LG,stat.ML,cs.AI},
pdfauthor={Fabiola Espinosa Castellon,Eduardo Fernandes Montesuma,Fred Ngol\`e Mboula,Aur\'elien Mayoue,Antoine Souloumiac,C\'edric Gouy-Pallier},
pdfkeywords={Optimal Transport,Federated Learning,Domain Adaptation,Dictionary Learning},
}

\newacronym{ot}{OT}{Optimal Transport}
\newacronym{dp}{DP}{Differential Privacy}
\newacronym{fl}{FL}{Federated Learning}
\newacronym{erm}{ERM}{Empirical Risk Minimization}
\newacronym{ml}{ML}{Machine Learning}
\newacronym{dil}{DiL}{Dictionary Learning}
\newacronym{dd}{DD}{Dataset Distillation}
\newacronym{dc}{DC}{Dataset Condensation}
\newacronym{dm}{DM}{Distribution Matching}
\newacronym{mmd}{MMD}{Maximum Mean Discrepancy}
\newacronym{dadil}{DaDiL}{Dataset Dictionary Learning}
\newacronym{feddadil}{FedDaDiL}{Federated DaDiL}
\newacronym{spc}{SPC}{Samples Per Class}
\newacronym{ipm}{IPM}{Integral Probability Metric}
\newacronym{svm}{SVM}{Support Vector Machine}
\newacronym{da}{DA}{Domain Adaptation}
\newacronym{msda}{MSDA}{Multi-Source DA}
\newacronym{cstr}{CSTR}{Continuously Stirred Tank Reactor}
\newacronym{tep}{TEP}{Tennessee Eastmann Process}
\newacronym{har}{HAR}{Human Activity Recognition}
\newacronym{wbt}{WBT}{Wasserstein Barycenter Transport}
\newacronym{co}{CO}{Caltech-Office 10}
\newacronym{cwru}{CWRU}{Case Western Reserve University}
\newacronym{sota}{SOTA}{State-of-the-Art}
\newacronym{wjdot}{Weighted Joint Distribution OT}{WJDOT}
\newacronym{sgd}{SGD}{Stochastic Gradient Descent}
\newacronym{noniid}{non-IID}{non-independent and non-identically distributed}
\newacronym{fedmsda}{Fed-MSDA}{Federated MSDA}
\newacronym{fedda}{FedDA}{Federated DA}

\renewcommand{\min}[1]{\underset{#1}{\text{min}}\,}

\renewcommand{\inf}[1]{\underset{#1}{\text{inf}}\,}

\newcommand{\iid}[0]{\overset{i.i.d.}{\sim}}

\newcommand{\argmin}[1]{\underset{#1}{\text{argmin}}\,}

\definecolor{myorange}{rgb}{0.906,0.435,0.317}
\definecolor{myblue}{rgb}{0.0,0.314,0.408}
\definecolor{private}{rgb}{0.5,0.0,0.0}
\definecolor{public}{rgb}{0.17,0.62,0.17}

\theoremstyle{definition}

\newtheorem{definition}{Definition}

\tikzstyle{trapezium}=[fill=white, draw=black, shape=trapezium, rotate=-90, minimum height=1cm]
\tikzstyle{lossbox}=[fill={rgb,255: red,202; green,206; blue,255}, draw=black, shape=rectangle, minimum height=1.2cm, minimum width=1cm, align=center]
\tikzstyle{clfbox}=[fill=white, draw=black, shape=rectangle, minimum width=1cm, minimum height=1cm]
\tikzstyle{new style 2}=[fill=white, draw=black, shape=rectangle, align=center]
\tikzstyle{domainbox}=[fill=white, draw=black, shape=rectangle, minimum width=3cm, align=center]
\tikzstyle{longbox}=[fill=white, draw=black, shape=rectangle, minimum height=4cm, minimum width=1.2cm, align=center]
\tikzstyle{rotatednode}=[rotate=90]
\tikzstyle{circularnode}=[fill=none, draw=black, shape=circle]
\tikzstyle{blue_circle}=[fill={rgb,255: red,0; green,80; blue,104}, draw=none, shape=circle, minimum width=0.5cm]
\tikzstyle{orangecircle1}=[fill={rgb,255: red,231; green,111; blue,81}, draw=none, shape=circle, minimum width=0.5cm]
\tikzstyle{blue_square1}=[fill={rgb,255: red,0; green,80; blue,104}, draw=none, shape=rectangle, minimum width=0.5cm, minimum height=0.5cm]
\tikzstyle{blue_triangle1}=[fill={rgb,255: red,0; green,80; blue,104}, draw=none, shape=regular polygon, regular polygon sides=3]
\tikzstyle{widebox}=[fill=white, draw=black, shape=rectangle, minimum height=1.2cm, minimum width=10cm, align=center]
\tikzstyle{labeled domain}=[fill=none, draw={rgb,255: red,0; green,80; blue,104}, shape=circle, minimum width=1cm]
\tikzstyle{unlabeled domain}=[fill=none, draw={rgb,255: red,231; green,111; blue,81}, shape=circle, minimum width=1cm]
\tikzstyle{smallwidebox}=[fill=white, draw=black, shape=rectangle, minimum height=1.2cm, minimum width=5cm, align=center]

\tikzstyle{red edge}=[->, fill=none, draw={rgb,255: red,128; green,0; blue,0}]
\tikzstyle{blue edge}=[->, fill=none, draw={rgb,255: red,70; green,130; blue,180}]
\tikzstyle{green edge}=[->, fill=none, draw={rgb,255: red,44; green,160; blue,44}]
\tikzstyle{red dotted edge}=[->, dashed, fill=none, draw={rgb,255: red,128; green,0; blue,0}, thick]
\tikzstyle{blue dotted edge}=[->, dashed, fill=none, draw={rgb,255: red,70; green,130; blue,180}]
\tikzstyle{green dotted edge}=[->, dashed, fill=none, draw={rgb,255: red,44; green,160; blue,44}, thick]
\tikzstyle{red dotted line}=[-, fill=none, dashed, draw={rgb,255: red,128; green,0; blue,0}]
\tikzstyle{blue dotted line}=[-, fill=none, dashed, draw={rgb,255: red,70; green,130; blue,180}]
\tikzstyle{green dotted line}=[-, fill=none, dashed, draw={rgb,255: red,44; green,160; blue,44}]
\tikzstyle{black edge}=[->]
\tikzstyle{black dashed line}=[-, dashed]
\tikzstyle{thick black arrow}=[->, thick]
\tikzstyle{thick black edge}=[-, thick]
\tikzstyle{thick black dotted line}=[->, thick, dashed]
\tikzstyle{semi transparent dashed black line}=[-, opacity=0.2, dashed]

\begin{document}
\maketitle

\def\thefootnote{*}\footnotetext{These authors contributed equally to this work}

\begin{abstract}
In this article, we propose an approach for federated domain adaptation, a setting where distributional shift exists among clients and some have unlabeled data. The proposed framework, FedDaDiL, tackles the resulting challenge through dictionary learning of empirical distributions. In our setting, clients' distributions represent particular domains, and FedDaDiL collectively trains a federated dictionary of empirical distributions. In particular, we build upon the Dataset Dictionary Learning framework by designing collaborative communication protocols and aggregation operations. The chosen protocols keep clients' data private, thus enhancing overall privacy compared to its centralized counterpart. We empirically demonstrate that our approach successfully generates labeled data on the target domain with extensive experiments on (i) Caltech-Office, (ii) TEP, and (iii) CWRU benchmarks. Furthermore, we compare our method to its centralized counterpart and other benchmarks in federated domain adaptation.
\end{abstract}


\section{Introduction}\label{sec:intro}

In \gls{ml}, \gls{fl} has emerged as a training paradigm for distributing the training of large scale models over a network of devices~\cite{mcmahan2017communication}. Nonetheless, this paradigm is substantially challenged when the clients, i.e., the data holders, follow different probability distributions, a case known as \emph{non-i.i.d.} \gls{fl}.

Further, in \gls{da}, a sub-problem within transfer learning~\cite{pan2009survey}, a labeled source domain adapts towards an unlabeled target domain. These domains are characterized by their different probability distributions. \gls{msda}, a later generalization, considers multiple heterogeneous source distributions~\cite{crammer2008learning}. Naturally, there is a parallel between \gls{da} and non-i.i.d. \gls{fl}, as both learning problems deal with distributional shifts. In this paper we propose to view each domain in \gls{msda} as a client in \gls{fl}.

As a solution to client heterogeneity in \gls{msda}, we propose to adapt the \gls{dadil} framework of~\cite{montesuma2023learning} to a federated setting. Especially, we show that the \gls{dadil} variables, i.e., the atoms and barycentric coordinates, are decoupled, so that training is done without the need for clients to communicate their data between themselves.

Section~\ref{sec:background} presents the overall background to our method. Section~\ref{sec:methodology} presents our main contribution, i.e. \gls{feddadil}. Section~\ref{sec:experiments} presents and discusses our experiments, and finally section~\ref{sec:conclusion} concludes this paper.

\section{Background}\label{sec:background}

\subsection{Optimal Transport}

\gls{ot} measures the minimum cost of transporting one distribution to another. We use the discrete Kantorovich formulation of \gls{ot}~\cite[Chapter 3]{peyre2017computational}. We refer readers to~\cite{montesuma2023recent} for further insight of \gls{ot} for \gls{ml}, including \gls{da}. 
Let $\mathbf{x}_{i}^{(Q)} \iid Q$, $i=1,\cdots, n$. We approximate $Q$ empirically,
\begin{align*}
    \hat{Q}(\mathbf{x}) &= \dfrac{1}{n}\sum_{i=1}^{n}\delta(\mathbf{x}-\mathbf{x}_{i}^{(Q)}).
\end{align*}
Given distributions $P$ and $Q$, with $m$ and $n$ samples each, let $\Pi(P, Q) = \{\pi\in\mathbb{R}^{n\times m}_{+}:\pi\mathbf{1}_{m}=n^{-1},\pi^{T}\mathbf{1}_{n}=m^{-1}\}$. Matrices $\pi \in \Pi(P,Q)$ are called transport plans between $P$ and $Q$. Let $C_{ij} = c(\mathbf{x}_{i}^{(P)},\mathbf{x}_{j}^{(Q)})$ be the ground-cost matrix,
\begin{align*}
    \pi^{\star} &= \argmin{\pi\in\Pi(P,Q)}\langle \pi, \mathbf{C} \rangle_{F},
\end{align*}
is called \gls{ot} plan. When $c$ is a distance between samples, the quantity $W_{c}(\hat{P},\hat{Q}) = \langle \pi^{\star},\mathbf{C} \rangle_{F}$ is called the Wasserstein distance between distributions $\hat{P}$ and $\hat{Q}$. As a distance, it allows for the definition of a barycenter of distributions~\cite{agueh2011barycenters}.
\begin{definition}\label{def:wbary}
For distributions $\mathcal{P} = \{P_{k}\}_{k=1}^{K}$ and weights $\alpha \in \Delta_{K}$, the Wasserstein barycenter is a solution to,
\begin{align*}
    B^{\star} = \mathcal{B}(\alpha;\mathcal{P}) = \inf{B}\sum_{k=1}^{K}\alpha_{k}W_{c}(P_{k}, B).
\end{align*}
Henceforth we call $\mathcal{B}(\cdot;\mathcal{P})$ barycentric operator.
\end{definition}

\subsection{Federated Learning}

In classification, we seek a function $h:\mathcal{X}\rightarrow\{1,\cdots,n_{c}\}$, where $n_{c}$ is the number of classes, and $\mathcal{X}$ is called input space (e.g., images, signals). In centralized supervised learning, a classifier is fit via empirical risk minimization~\cite{vapnik2013nature},
\begin{align}
    h^{\star} = \argmin{h\in\mathcal{H}}f(h):=\dfrac{1}{n}\sum_{i=1}^{n}\mathcal{L}(h(\mathbf{x}_{i}^{(Q)}), y_{i}^{(Q)}),\label{eq:erm}
\end{align}
for $\mathbf{x}_{i}^{(Q)} \iid Q$ and $y_{i}^{(Q)} = h_{0}(\mathbf{x}_{i}^{(Q)})$. For neural nets, $h$ is parametrized via the neural net's weights $\theta$, which is found via gradient descent on $f(\theta) = f(h_{\theta})$.

\gls{fl}~\cite{mcmahan2017communication} distributes the minimization of eq.~\ref{eq:erm} on a set of $N$ clients,
\begin{equation}
\min{\theta}{\sum_{\ell=1}^{N}{w_{\ell}}f_{\ell}(\theta)}\text{ where, }w_{\ell}=\frac{n_{\ell}}{\sum_{\ell=1}^{N}{n_{\ell}}},\label{equ:fed_erm}
\end{equation}
where $f_{\ell}(\theta)$ is the loss (eq.~\ref{eq:erm}) of client $\ell$. This setting  was first formalized by the algorithm FedAVG~\cite{mcmahan2017communication}, in which a server averages clients’ models at each round of communication. Yet, in real-world scenarios clients generate opinion-based and personal data from different environments. Hence, the massively distributed data can be non-i.i.d. between different clients participating in the global training. As a result, the global performance of a federated model is degraded~\cite{zhao2018federated}. Numerous methods seek to tackle client heterogeneity through the perspective of model parameters or optimization trajectory~\cite{karimireddy2020scaffold,liu2020secure}. However, these methods assume that all clients can access labeled data, which might not hold in practice.

\subsection{Domain Adaptation}

\gls{da} is a sub-problem within transfer learning~\cite{pan2009survey}. In this context, a domain is a pair $(\mathcal{X},Q)$ of a feature space and a feature marginal distribution. In \gls{da}, one has different but related domains over the same feature space. The goal is to adapt models learned on a \emph{labeled source distribution} $Q_{S}$ to an \emph{unlabeled target distribution} $Q_{T}$. We focus on \gls{msda}~\cite{crammer2008learning}, which tackles the learning problem under heterogeneous source distributions.
Data comes from multiple distributions $\mathcal{Q}_{S} = \{\hat{Q}_{S_{\ell}}\}_{\ell=1}^{N_{S}}$, where $\hat{Q}_{S_{i}} \neq Q_{S_{j}}$ for $i \neq j$.

\gls{ot} previously contributed to \gls{msda} in various ways. For instance,~\cite{montesuma2021icassp,montesuma2021cvpr} use Wasserstein barycenters for aggregating the heterogeneous sources into a common distribution, whereas~\cite{montesuma2023learning} proposed a novel method unifying \gls{msda} and \gls{dil} through Wasserstein barycenters. However, none of these methods consider a federated adaptation setting. In this sense, other \gls{msda} frameworks were considered for \gls{fedda}. For instance,~\cite{peng2019federated} employs feature disentanglement and adversarial training for aligning source clients with target data held by a server. In addition,~\cite{feng2021kd3a} employs knowledge distillation on models learned privately on source domain data, while reducing the amount of communication rounds needed for convergence.

\section{Methodology}\label{sec:methodology}

\begin{figure}[ht]
    \centering
    \resizebox{\linewidth}{!}{\begin{tikzpicture}
	\begin{pgfonlayer}{nodelayer}
            \node [style=none, align=center] (11) at (-9, 8) {Source Client$_{1}$};
		\node [style=labeled domain] (1) at (-9, 6) {\Large{\faMobile}};
		\node [style=labeled domain] (2) at (0, 1) {\Large{\faMobile}};
            \node [style=none, align=center] (33) at (3.5, 1) {Source Client$_{2}$};
		\node [style=unlabeled domain] (3) at (9, 6) {\Large{\faMobile}};
            \node [style=none, align=center] (33) at (9, 8) {Target Client$_{1}$};
		\node [style=none] (4) at (-4, 7.5) {};
		\node [style=none] (5) at (-4, 4.5) {};
		\node [style=none] (6) at (4, 7.5) {};
		\node [style=none] (7) at (4, 4.5) {};
		\node [style=labeled domain] (8) at (-2.5, 6) {$\hat{P}_{1}$};
		\node [style=labeled domain] (9) at (0, 6) {$\hat{P}_{2}$};
		\node [style=labeled domain] (10) at (2.5, 6) {$\hat{P}_{2}$};
		\node [style=none] (11) at (4, 6.5) {};
		\node [style=none] (12) at (4, 5.5) {};
		\node [style=none] (13) at (0.5, 4.5) {};
		\node [style=none] (14) at (-0.5, 4.5) {};
		\node [style=none] (15) at (-4, 5.5) {};
		\node [style=none] (16) at (-4, 6.5) {};
		\node [style=none] (18) at (9, 2.5) {\faLock};
            \node [style=none,align=center] (181) at (-9, 2.3) {$\textcolor{private}{\alpha_{1}^{(e+1)}} \leftarrow \textcolor{private}{\alpha_{1}^{(e+1)}} - \eta\textcolor{private}{\nicefrac{\partial f_{\ell}}{\partial \alpha_{1}}}$};
		\node [style=none] (19) at (0, -2.5) {\faLock};
            \node [style=none] (192) at (0, 8) {Server \faServer};
            \node [style=none,align=center] (191) at (-6.2, 4) {$\textcolor{public}{\mathcal{P}_{g}^{(r)}}$};
		\node [style=none,align=center] (191) at (-6.2, 8) {$\textcolor{public}{\mathcal{P}_{1}^{\star}}$};
            \node [style=none] (20) at (6.25, 4) {\faUnlock};
	\end{pgfonlayer}
	\begin{pgfonlayer}{edgelayer}
		\draw [style=red dotted edge, in=-120, out=-60, loop] (3) to ();
		\draw [style=red dotted edge, in=-120, out=-60, loop] (2) to ();
		\draw [style=red dotted edge, in=-120, out=-60, loop] (1) to ();
		\draw (5.center) to (4.center);
		\draw (4.center) to (6.center);
		\draw (6.center) to (7.center);
		\draw (7.center) to (5.center);
		\draw [style=green dotted edge, bend right] (3) to (11.center);
		\draw [style=green dotted edge, bend right] (12.center) to (3);
		\draw [style=green dotted edge, bend right] (2) to (13.center);
		\draw [style=green dotted edge, bend right] (14.center) to (2);
		\draw [style=green dotted edge, bend left] (1) to (16.center);
		\draw [style=green dotted edge, bend right=330] (15.center) to (1);
	\end{pgfonlayer}
\end{tikzpicture}}
    \caption{\gls{feddadil}. A server holds public (\faUnlock) \textcolor{public}{atoms}. It sends a copy to each client, which performs local updates. During these updates, clients' \textcolor{private}{barycentric coordinates} are kept private (\faLock). The learning is done collaboratively without the need to communicate clients' data directly.}
    \label{fig:feddadil}
\end{figure}
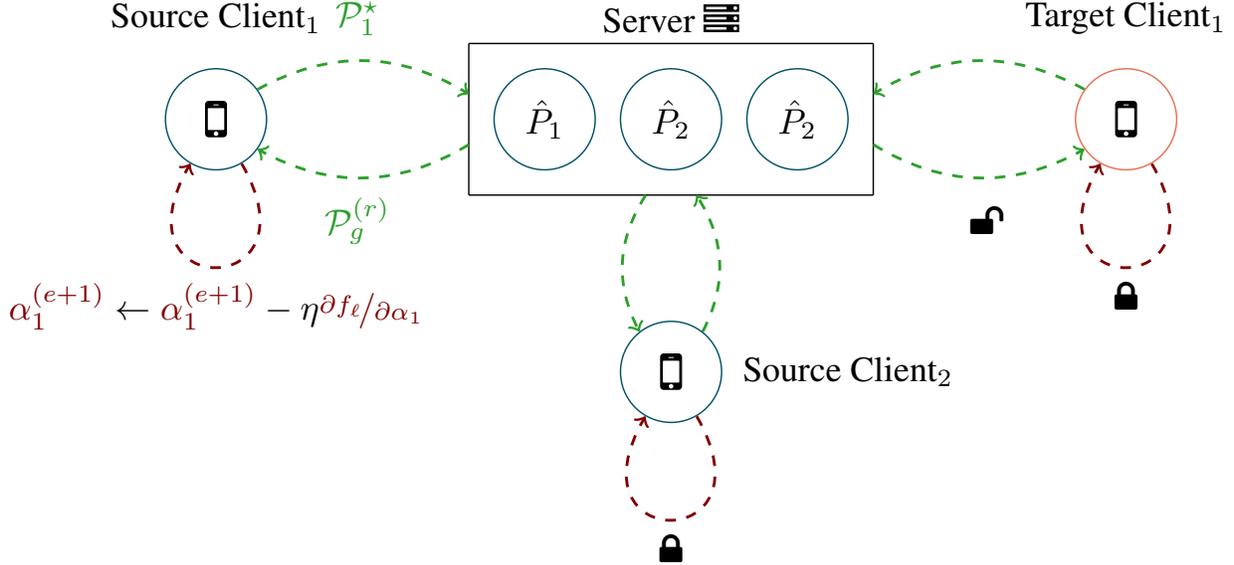

\noindent\textbf{Problem Statement.} In the following, we assume that each domain in \gls{msda} is a client in \gls{fl}.
This setting puts our algorithm in the non-i.i.d. scenario~\cite{zhao2018federated}. Among the existing clients, one client, called \emph{target}, has \textcolor{myorange}{unlabeled data}. Our goal is to provide \textcolor{myblue}{labeled data} that follows this client's probability distribution, so that a classifier can reliably predict on such domain. Our proposed \gls{feddadil} learns a dictionary of \textcolor{public}{public} empirical distributions $\mathcal{P}$, called atoms, and \textcolor{private}{private} barycentric coordinates $\mathcal{A}$. We illustrate this idea in figure~\ref{fig:feddadil}.

\subsection{Centralized Dataset Dictionary Learning}

In~\cite{montesuma2023learning}, authors proposed a novel \gls{dil} framework over empirical distributions. Given a set of datasets $\mathcal{Q} = \{\hat{Q}_{S_{\ell}}\}_{\ell=1}^{N_{S}} \cup \{\hat{Q}_{T}\}$, \gls{dadil} learns a set of atoms $\mathcal{P} = \{\hat{P}_{k}\}_{k=1}^{K}$, and barycentric coordinates $\mathcal{A} = \{\alpha_{\ell}:\alpha_{\ell} \in \Delta_{K}\}_{\ell=1}^{N}$, such that,
\begin{align}
(\mathcal{P}^{\star},\mathcal{A}^{\star}) = \argmin{\mathcal{P},\mathcal{A}}f(\mathcal{P},\mathcal{A}) := \dfrac{1}{N}\sum_{\ell=1}^{N}f_{\ell}(\alpha_{\ell},\mathcal{P}),\label{eq:dadil}
\end{align}
where $N=N_{S}+1$, and $f_{\ell}$ is the objective function of each domain, that is,
\begin{equation}
    f_{\ell}(\alpha_{\ell},\mathcal{P}) = \begin{cases}
        W_{c}(\hat{Q}_{\ell},\mathcal{B}(\alpha_{\ell};\mathcal{P}))& 
\hat{Q}_{\ell} \text{ is labeled},\\
        W_{2}(\hat{Q}_{\ell},\mathcal{B}(\alpha_{\ell};\mathcal{P}))&\text{ otherwise},\\
    \end{cases}\label{eq:local loss}
\end{equation}

In practice, the minimization of eq.~\ref{eq:dadil} is done w.r.t. $\{(\mathbf{X}^{(P_{k})}, \mathbf{Y}^{(P_{k})})\}_{k=1}^{K}$, and the barycentric coordinates $\alpha_{\ell}$ of each domain. As shown in~\cite[Alg. 2]{montesuma2023learning}, eq.~\ref{eq:dadil} is optimized through \gls{sgd}.

\subsection{Federated Dataset Dictionary Learning}

\noindent\textbf{FedDaDiL (Algorithm 1).} On one hand, \gls{feddadil} follows~\cite[Alg. 2]{montesuma2023learning} in using \gls{sgd}. On the other hand, the optimization of $(\mathcal{P},\mathcal{A})$ is distributed among the clients. \gls{feddadil} works as follows. (1) The \emph{server} initializes a global dictionary $\mathcal{P}_{g}^{(0)}$. (2) At round $r$, the \emph{server} communicates a mini-batch from atoms to each client. (3) The \emph{client} clones this mini-batch into a local version, i.e., $\mathcal{P}_{\ell}^{(0)} \leftarrow \mathcal{P}_{g}^{(r)}$, and optimizes $\mathcal{P}_{\ell}$, as well as its own barycentric coordinates $\alpha_{\ell}$ through $E$ local epochs (cf. algorithm~\ref{alg:client_update}). Due to the decoupling introduced by \gls{feddadil}, the client's barycentric coordinates can \textcolor{private}{\textbf{remain private}}. Finally, at the end of the communication round, the server aggregates the different dictionary versions generated by the clients.

\noindent\textbf{Updating Client Parameters (Algorithm 2).} Analogously to \emph{FedAVG}~\cite{mcmahan2017communication}, each client participating in \gls{feddadil} needs to locally update parameters w.r.t. its own data. This is described in algorithm~\ref{alg:client_update}. In a nutshell the client calculates its own loss $f_{\ell}$ with mini-batches, and updates $\mathcal{P}_{\ell}$ and $\alpha_{\ell}$ internally via gradient descent. The barycentric coordinates $\alpha_{\ell}$ need to be projected into the simplex, as in~\cite[Sec. 4.2]{montesuma2023learning}.

\begin{minipage}{0.55\textwidth}
\begin{algorithm}[H]
\caption{\gls{feddadil}. The $N$ Clients are indexed by $\ell$. $n_{b}$ is the batch size.}
\begin{algorithmic}[1]
    \small
    \STATE Server initializes $\mathcal{P}_{g}^{(0)} = \{\hat{P}_{k}^{(0)}\}_{k=1}^{K}$
    \STATE client$_{\ell}$ initializes $\alpha_{\ell}^{(0)} \in \Delta_{K}$, $\forall \ell=1\cdots,N_{c}$
    \FOR{each round $r=1\cdots,R$}
        \STATE Sample clients $\mathcal{C} \subset \{1,\cdots,N\}$
        \STATE Communicate $\mathcal{P}_{g}^{(r)} = \{\{(\mathbf{x}_{i}^{(P_{k})},\mathbf{y}_{i}^{(P_{k})})\}_{i=1}^{n_{b}}\}_{k=1}^{K}$, $\forall\ell \in \mathcal{C}$
        \FOR{client $\ell \in \mathcal{C}$}
            \STATE Initialize local dictionary $\mathcal{P}_{\ell}^{(0)} \leftarrow \mathcal{P}_{g}^{(r)}$
            \STATE $\mathcal{P}_{\ell}^{(r)} = \text{ClientUpdate}(\mathcal{P}_{\ell}^{(0)},\alpha_{\ell}^{(r)})$
            \STATE client$_{\ell}$ sends $\mathcal{P}_{\ell}^{(r)}$ to server.
        \ENDFOR
        \STATE $\mathcal{P}^{(r+1)}_{g} \leftarrow \text{ServerAggregate}(\{\mathcal{P}_{\ell}^{(r)}\}_{\ell \in \mathcal{C}})$
    \ENDFOR
    \ENSURE Dictionary $\mathcal{P}^{\star}$ and weights $\mathcal{A}^{\star}$.
\end{algorithmic}
\label{alg:fed_dadil}
\end{algorithm}
\end{minipage}
\hfill
\begin{minipage}{0.4\textwidth}
\begin{algorithm}[H]
\caption{ClientUpdate.}
\begin{algorithmic}[1]
    \small
    \REQUIRE Local atom $\mathcal{P}$. Set of weights $\alpha_{\ell} \in \Delta_{K}$. $E$ is the number of local epochs. Learning rate $\eta$.
    \FOR{local epoch $e=1,\cdots,E$}
        \FOR{batch $b=1,\cdots,B$}
            \STATE $f_{\ell}(\alpha_{\ell};\mathcal{P}_{\ell}^{(e)}) = W_{c}(\hat{Q}_{\ell},\mathcal{B}(\alpha_{\ell};\mathcal{P}_{\ell}^{(e)}))$
            \STATE $\mathbf{x}_{i}^{(P_{k})} \leftarrow \mathbf{x}_{i}^{(P_{k})} - \eta \nicefrac{\partial f_{\ell}}{\partial \mathbf{x}_{i}^{(P_{k})}}(\alpha_{\ell},\mathcal{P})$
            \STATE $\mathbf{y}_{i}^{(P_{k})} \leftarrow \mathbf{y}_{i}^{(P_{k})} - \eta \nicefrac{\partial f_{\ell}}{\partial \mathbf{y}_{i}^{(P_{k})}}(\alpha_{\ell},\mathcal{P})$
            \STATE $\alpha_{\ell} \leftarrow \text{proj}_{\Delta_{K}}(\alpha_{\ell} - \eta \nicefrac{\partial f_{\ell}}{\partial \alpha_{\ell}}(\alpha_{\ell},\mathcal{P}))$
        \ENDFOR
    \ENDFOR
    \STATE Client sets $\alpha_{\ell}^{(r+1)} \leftarrow \alpha_{\ell}^{\star}$.
    \ENSURE $\mathcal{P}_{\ell}^{\star}$.
\end{algorithmic}
\label{alg:client_update}
\end{algorithm}
\end{minipage}

\noindent\textbf{Aggregating Dictionary Versions.} The last ingredient in \gls{feddadil} is aggregating the multiple versions $\{\mathcal{P}_{\ell}^{(r)}\}_{\ell\in\mathcal{C}}$. Overall, equation~\ref{eq:dadil} includes heterogeneous terms. For instance, the target domain objective does not include labels. For that reason, averaging dictionary versions in analogy with \emph{FedAVG} is not feasible. Therefore, we seek to mimic the reasoning behind \gls{dadil}, who optimize a different mini-batch of atoms for each domain. As a consequence, we propose to randomly select one version from the pool of client's local dictionaries. As we show in our experiments, this still yields good solutions for \gls{dadil}.

\noindent\textbf{Complexity.} As in \gls{dadil}, the cost of processing a batch is proportional to \gls{ot}'s complexity, i.e., $\mathcal{O}(n_{b}^{3}\log n_{b})$. The complexity ultimately depends on the number of atoms $K$, as analyzed by~\cite{montesuma2023learning}. In terms of communication, \gls{feddadil} is lightweight. Indeed, it only communicates mini-batches. For comparison, assume a reasonably large batch size of $n_{b}=128$, and $d=2048$. This implies communicating $n_{b}\times d = 2^{18}$ bytes of information, which is considerably smaller than modern deep neural nets.

\section{Experiments and Discussion}\label{sec:experiments}

We evaluate \gls{feddadil} on 3 benchmarks: (1) \gls{tep}~\cite{montesuma2023multi}, (2) \gls{cwru} and (3) Caltech-Office 10~\cite{gong2012geodesic}. While the first two are concerned with fault diagnosis from time series, the third is a standard benchmark in visual \gls{da}. \gls{tep} consists of signals coming from the simulation of a large-scale complex chemical process. We reproduce the experimental setting of~\cite{montesuma2023multi}, which includes deep \gls{da} methods. \gls{cwru} consists of vibration signals collected from rotating machinery, for which, we reproduce the experimental setting of~\cite{montesuma2023learning}. For \gls{tep} and \gls{cwru}, we use features extracted from a neural network trained with source-domain data. For Caltech-Office, we use convolutional features of a DeCAF netowrk~\cite{donahue2014decaf}.

\begin{table}[ht]
    \centering
    \caption{Overview of benchmarks used in our experiments.}
    \begin{tabular}{lllll}
         \toprule
         Benchmark & \# Samples & \# Domains & \# Classes & \# Features  \\
         \midrule
         TEP & 17289 & 6 & 29 & 128\\
         CWRU & 24000 & 3 & 10 & 256\\
         Caltech-Office 10 & 2533 & 4 & 10 & 4096\\
         \bottomrule
    \end{tabular}
    \label{tab:datasets}
\end{table}

We compare \gls{feddadil} with centralized \gls{dadil}~\cite{montesuma2023learning}. For completeness, we report \gls{wjdot}~\cite{turrisi2022multi} and \gls{wbt}~\cite{montesuma2021cvpr} on the Caltech-Office 10 benchmark. Our main point of comparison are other federated baselines: (1) \emph{FedAVG}~\cite{mcmahan2017communication}, which is the federated implementation of the Source-Only baseline. On \gls{tep} and \gls{cwru}, we also compare with (2) \emph{FedMMD}, (3) \emph{FedDANN}, (4) \emph{FedWDGRL}, and (5) KD3A~\cite{feng2021kd3a}. (2-4) are adaptations of the methods of~\cite{long2015learning,ganin2016domain,shen2018wasserstein}. We also considered FADA~\cite{peng2019federated}, but their neural net architecture could not be directly adapted to time series data.

\subsection{Dictionary Learning}\label{sec:dil}

\begin{figure}[ht]
    \centering
\begin{subfigure}{0.4\linewidth}
    \includegraphics[width=\linewidth]{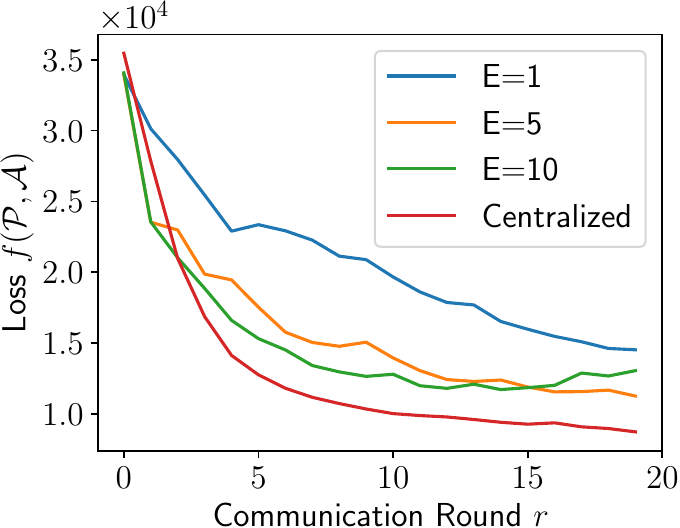}
    \caption{}
    \label{fig:local_it_dil}
\end{subfigure}\hspace{10pt}
\begin{subfigure}{0.4\linewidth}
    \includegraphics[width=\linewidth]{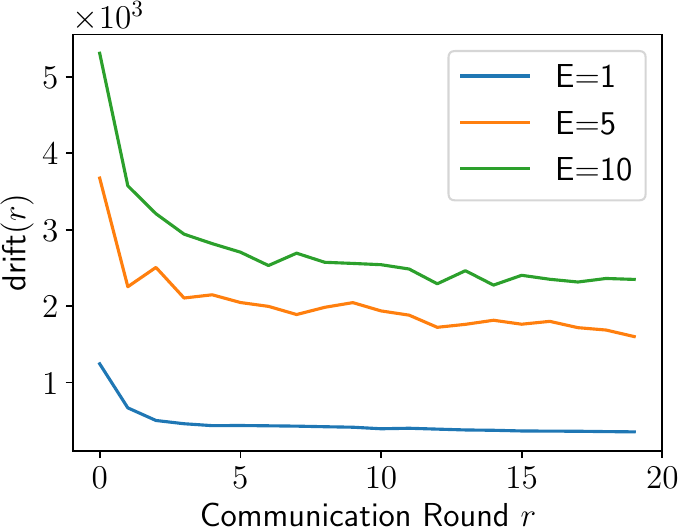}
    \caption{}
    \label{fig:client_drift}
\end{subfigure}
    \caption{(a) \gls{dil} loss (eq.~\ref{eq:dadil}) and (b) client drift (eq.~\ref{eq:drift}) as a function of communication rounds.}
    \label{fig:dil_analysis}
\end{figure}

We probe the effect of local parameter $E \in \{1, 5, 10\}$ during \gls{feddadil} optimization, and in comparison with centralized \gls{dadil} for the Caltech-Office benchmark  (figure~\ref{fig:local_it_dil}). Globally, the centralized version achieves a better minimum w.r.t. the federated version. 
Moreover, since we randomly select a client's atom version at each round,  figure~\ref{fig:local_it_dil} implies that individual clients find atoms that generalize to other distributions. The \gls{dil} loss thus decreases with increasing $E$ because the chosen client performs more mini-batch gradient steps, leading to a better estimate of a local minimum of eq.~\ref{eq:dadil}.

Next, we evaluate if clients' atom versions drift apart during optimization, or converge towards a possible solution. At round $r$, client $\ell$ has version $\mathcal{P}_{\ell}^{(r)} = \{\hat{P}_{\ell,k}^{(r)}\}_{k=1}^{K}$. The drift is,
\begin{align}
    \text{drift}(r) = \sum_{\ell_{1} \neq \ell_{2}}\dfrac{1}{K}\sum_{k=1}^{K}W_{c}(\hat{P}_{\ell_{1},k}^{(r)},\hat{P}_{\ell_{2},k}^{(r)}).\label{eq:drift}
\end{align}

The drift between atom versions increases with $E$, implying that clients personalize their atoms, but it decreases with the number of rounds as convergence is reached (figure~\ref{fig:client_drift}), indicating that optimization converges to a local minimum.

\subsection{Federated Domain Adaptation}\label{sec:msda}

\begin{table}[ht]
\caption{Experimental results on 3 \gls{msda} benchmarks. Tables show mean $\pm$ standard deviation for 5 independent runs on each experiment. $^{\dagger}$, $^{\ddag}$, $^{\mathsection}$ and $^{\star}$ denote results from~\cite{montesuma2021cvpr,turrisi2022multi,montesuma2023multi} and~\cite{montesuma2023learning}, receptively.}
\centering
\begin{subtable}[t]{0.60\linewidth}
    \caption{Tennessee Eastman Process.}
    \resizebox{\linewidth}{!}{
    \begin{tabular}{lccccccc}
        \toprule
        Target & 1 & 2 & 3 & 4 & 5 & 6 & Avg. \\
        \midrule
        Source-Only$^{\mathsection}$ & 80.82 $\pm$ 0.96 & 63.69 $\pm$ 1.71 & 87.47 $\pm$ 0.99 & 79.96 $\pm$ 1.07 & 74.40 $\pm$ 1.53 & 84.53 $\pm$ 1.12 & 78.47\\
        MMD$^{\mathsection}$ & 82.86 $\pm$ 2.00 & 65.31 $\pm$ 3.42 & 80.51 $\pm$ 2.76 & 79.79 $\pm$ 2.53 & 69.09 $\pm$ 3.98 & 80.04 $\pm$ 3.08 & 76.33\\
        DANN$^{\mathsection}$ & 52.72 $\pm$ 2.40 & 56.13 $\pm$ 4.10 & 48.88 $\pm$ 2.83 & 62.89 $\pm$ 3.14 & 56.43 $\pm$ 2.33 & 43.28 $\pm$ 2.12 & 53.39\\
        WDGRL & 44.34 $\pm$ 7.46 & 55.12 $\pm$ 5.64 & 49.02 $\pm$ 5.94 & 72.39 $\pm$ 4.65 & 70.95 $\pm$ 5.16 & 44.01 $\pm$ 3.93 & 55.97\\
        DaDiL-R$^{\mathsection}$ & 91.97 $\pm$ 1.22 & 77.15 $\pm$ 1.32 & 85.41 $\pm$ 1.69 & 89.39 $\pm$ 1.03 & 84.49 $\pm$ 1.95 & 88.44 $\pm$ 1.29 & 86.14\\
        DaDiL-E$^{\mathsection}$ & 90.45 $\pm$ 1.02 & 77.08 $\pm$ 1.21 & 86.79 $\pm$ 2.14 & 89.01 $\pm$ 1.35 & 84.04 $\pm$ 3.16 & 87.85 $\pm$ 1.06 &  85.87\\
        \midrule
        FedAVG & 85.28 $\pm$ 1.37 & 47.45 $\pm$ 2.56 & 75.72 $\pm$ 1.36 & 73.33 $\pm$ 1.91 & 67.36 $\pm$ 2.40 & 72.52 $\pm$ 1.97 & 72.52\\
        FedMMD & 62.13 $\pm$ 1.12 & 50.48 $\pm$ 3.22 & 56.26 $\pm$ 1.60 & 50.47 $\pm$ 3.63 & 51.19 $\pm$ 4.51 & 57.05 $\pm$ 2.92 & 54.59\\
        FedDANN & 60.93 $\pm$ 2.42 & 52.47 $\pm$ 5.55 & 59.19 $\pm$ 1.90 & 52.91 $\pm$ 4.52 & 54.63 $\pm$ 2.14 & 57.09 $\pm$ 3.23 & 56.20\\
        FedWDGRL & 82.65 $\pm$ 1.91 & 59.54 $\pm$ 2.71 & 82.22 $\pm$ 2.08 & 75.28 $\pm$ 1.32 & 72.97 $\pm$ 5.63 & 77.70 $\pm$ 2.89 & 75.06\\
        KD3A & 72.52 $\pm$ 3.04 & 18.96 $\pm$ 4.54 & 81.02 $\pm$ 2.40 & 74.42 $\pm$ 1.60 & 67.18 $\pm$ 2.37 & 78.22 $\pm$ 2.14 & 65.38\\
        FedDaDiL-R & \textbf{87.41} $\pm$ \textbf{0.72} & \underline{73.63} $\pm$ \underline{2.46} & \underline{87.41} $\pm$ \underline{2.86} & \textbf{88.06} $\pm$ \textbf{0.79} & \underline{81.99} $\pm$ \underline{3.39} & \textbf{86.15} $\pm$ \textbf{1.46} & \underline{84.11}\\
        FedDaDiL-E & \underline{87.31} $\pm$ \underline{0.71} & \textbf{73.81} $\pm$ \textbf{2.73} & \textbf{87.5}4 $\pm$ \textbf{2.46} & \underline{87.88} $\pm$ \underline{0.56} & \textbf{82.44} $\pm$ \textbf{3.25} & \textbf{86.05} $\pm$ \textbf{1.00} & \textbf{84.17}\\
        \bottomrule
    \end{tabular}
    }
\end{subtable}\hfill
\begin{subtable}[t]{0.38\linewidth}
    \caption{CWRU.}
    \resizebox{\linewidth}{!}{
    \begin{tabular}{lcccc}
        \toprule
        Target & A & B & C & Avg. \\
        \midrule
        Source-Only$^{\star}$ & 70.90 $\pm$ 0.40 & 79.76 $\pm$ 0.11 & 72.26 $\pm$ 0.23 & 74.31 \\
        MMD & 84.29 $\pm$ 3.76 & 73.34 $\pm$ 3.87 & 79.08 $\pm$ 2.26 & 78.90\\
        DANN & 78.55 $\pm$ 5.38 & 64.78 $\pm$ 5.90 & 52.37 $\pm$ 5.40 & 65.24\\
        WDGRL & 84.24 $\pm$ 5.20 & 76.89 $\pm$ 4.27 & 76.98 $\pm$ 2.49 & 79.37\\
        DaDiL-R$^{\star}$ & {99.86} $\pm$ {0.21} & {99.85} $\pm$ {0.08} & {100.00} $\pm$ {0.00}  & {99.90}\\
        DaDiL-E$^{\star}$ & {93.71} $\pm$ {6.50} & {83.63} $\pm$ {4.98} & {99.97} $\pm$ {0.05}  & {92.33} \\
        \midrule
        FedAVG & 67.34 $\pm$ 0.36 & 62.23 $\pm$ 0.23 & 68.34 $\pm$ 0.29 & 65.97\\
        FedMMD & 79.58 $\pm$ 3.49 & 76.11 $\pm$ 2.00 & 75.22 $\pm$ 1.64 & 76.97\\
        FedDANN & 80.32 $\pm$ 7.77 & 75.82 $\pm$ 4.01 & \underline{80.13} $\pm$ \underline{2.32} & \underline{78.76}\\
        FedWDGRL & 79.23 $\pm$ 4.46 & 72.94 $\pm$ 5.10 & 76.16 $\pm$ 1.85 & 76.11\\
        KD3A & \underline{81.02} $\pm$ \underline{2.92} & \underline{78.04} $\pm$ \underline{4.05} & 74.64 $\pm$ 5.65 & 77.90 \\
        FedDaDiL-R & \textbf{99.65} $\pm$ \textbf{0.35} & \textbf{94.23} $\pm$ \textbf{5.11} & \textbf{100.00} $\pm$ \textbf{0.00} & \textbf{97.96}\\
        FedDaDiL-E & \textbf{99.60} $\pm$ \textbf{0.44} & \textbf{94.26} $\pm$ \textbf{4.87} & \textbf{100.00} $\pm$ \textbf{0.00} & \textbf{97.95}\\
        \bottomrule
    \end{tabular}
    }
\end{subtable}\\
\begin{subtable}[t]{0.45\linewidth}
    \caption{Caltech-Office 10.}
    \resizebox{\linewidth}{!}{
    \begin{tabular}{lccccc}
        \toprule
        Target & A & D & W & C & Avg. \\
        \midrule
        Source-Only$^{\dagger}$ & 90.55 $\pm$ 1.36 & 96.83 $\pm$ 1.33 & 88.36 $\pm$ 1.33 & 82.95 $\pm$ 1.26 & 89.67\\
        WJDOT$^{\ddag}$ & {94.23} $\pm$ {0.90} & {100.00} $\pm$ {0.00} & 89.33 $\pm$ 2.91 & 85.93 $\pm$ 2.07 & 92.37\\
        WBT$^{\dagger}$ & 92.74 $\pm$ 0.45 & 95.87 $\pm$ 1.43 & 96.57 $\pm$ 1.76 & 85.01 $\pm$ 0.84 & 92.55\\
        DaDiL-R$^{\star}$ & 94.06 $\pm$ 1.82 & {98.75} $\pm$ {1.71} & {98.98} $\pm$ {1.51} & {88.97} $\pm$ {1.06} & {95.19}\\
        DaDiL-E$^{\star}$ & {94.16} $\pm$ {1.58} & {100.00} $\pm$ {0.00} & {99.32} $\pm$ {0.93} & {89.15} $\pm$ {1.68} & {95.66}\\
        \midrule
        FedAVG & \textbf{93.95} $\pm$ \textbf{1.50} & 98.12 $\pm$ 1.71 & 92.54 $\pm$ 4.88 & 84.00 $\pm$ 1.74 & 92.15\\
        FedDaDiL-R & \underline{93.23} $\pm$ \underline{1.04} & \textbf{98.75} $\pm$ \textbf{2.79} & 97.96 $\pm$ 1.85 & \textbf{87.73} $\pm$ \textbf{1.02} & \textbf{94.42}\\
        FedDaDiL-E & 93.02 $\pm$ 0.94 & 98.12 $\pm$ 2.79 & \textbf{98.64} $\pm$ \textbf{1.42} & 86.84 $\pm$ 0.86 & \underline{94.16}\\
        \bottomrule
    \end{tabular}
    }
\end{subtable}
\label{tab:da_results}
\end{table}

The considered benchmarks, \gls{cwru}, Caltech-Office and \gls{tep}, pose an increasing challenge w.r.t. the federated setting. Indeed, these benchmarks have 3, 4, and 6 clients respectively. We present our results in table~\ref{tab:da_results}.

First, in \gls{fedda}, methods explicitly know from which domain data comes from. This is different from usual evaluation of single-source algorithms in the \gls{msda} setting, where algorithms concatenate source-domain data toghether. As such, some methods improve over their centralized baselines (e.g., DANN vs. FedDANN on \gls{cwru}). Nonetheless, for an increasing number of clients, and an increasing degree of heterogeneity, one can expect some performance degradation (e.g. FedMMD on \gls{tep}).

Second, \gls{feddadil} outperforms other \gls{fedda} methods. Indeed, a few aspects favor \gls{feddadil}: (i) our method does not need to update the neural net encoder, which is computationally complex. (ii) As \gls{dadil}, our proposed method interpolates distributional shift. As such, it is able to exploit regularities in the distributional shift between clients' distributions. Nonetheless, the federated setting is more challenging than centralized \gls{dadil}, as evidenced in Fig.~\ref{fig:local_it_dil}, and the performance degradation w.r.t. centralized \gls{dadil}.

\section{Conclusion}\label{sec:conclusion}

We present an innovative approach to address \gls{msda} within a \gls{fl} context. We introduce a federate strategy to train \gls{dadil}~\cite{montesuma2023learning}, without clients needing to directly communicate their data. As such, clients collaboratively build the atoms of a shared dictionary $\mathcal{P} = \{ {\hat{\mathcal{P}}_1,..., \hat{\mathcal{P}}_k} \}$, which models the distributional shift occurring between domains. Concurrently, clients also locally optimize weights that enable them to model their data distribution as an interpolation of atoms in the Wasserstein space. Clients' privacy is thus reinforced, as their weights are not shared.

In the \gls{fedda} setting, samples in the target domain are predicted either by using the Wasserstein barycenter of the target domain (FedDaDiL-R) or by using an ensemble of classifiers trained on the atoms (FedDadDiL-E).
Our experimental results improve classification performance by 11.60\%, 17.98\% and 2.27\% in signal, and image processing \gls{da} tasks. Further works includes analyzing our method through the lens of differential privacy, and extending our setting to the partitioning of domains throughout multiple clients.


\bibliographystyle{unsrt}
\bibliography{references}

\end{document}